\documentclass{article}
\usepackage[numbers,compress]{natbib}
\usepackage[final]{neurips_2024}

\usepackage[utf8]{inputenc} 
\usepackage[T1]{fontenc}    
\usepackage{hyperref}       
\usepackage{url}            
\usepackage{booktabs}       
\usepackage{amsfonts}       
\usepackage{nicefrac}       
\usepackage{microtype}      
\usepackage{xcolor}         
\usepackage{marvosym}

\usepackage{graphicx}
\usepackage{multirow}
\usepackage{amsmath}
\definecolor{cvprblue}{rgb}{0.21,0.49,0.74}
\usepackage{footnote}
\makesavenoteenv{tabular}
\makesavenoteenv{table}
\usepackage{pifont}

\newcommand{\sota}[1]{\textbf{#1}}
\newcommand{\subsota}[1]{\underline{#1}}
\definecolor{mgreen}{HTML}{D5E8D4}
\definecolor{mred}{HTML}{F8CECC}
\definecolor{airforceblue}{HTML}{5D8AA8}
\newcommand{\xmark}{\ding{55}}

\title{Generalizable Implicit Motion Modeling\\ for Video Frame Interpolation}

\author{%
    Zujin Guo, Wei Li$^{\ast}$, Chen Change Loy \\
    S-Lab, Nanyang Technological University\\
    \texttt{\{zujin.guo, wei.l, ccloy\}@ntu.edu.sg} \\
    {\tt\url{https://gseancdat.github.io/projects/GIMMVFI}}
}

\begin{document}

\maketitle
\let\thefootnote\relax\footnotetext{$^{\ast}$ Corresponding author}

\begin{abstract}
Motion modeling is critical in flow-based Video Frame Interpolation (VFI). Existing paradigms either consider linear combinations of bidirectional flows or directly predict bilateral flows for given timestamps without exploring favorable motion priors, thus lacking the capability of effectively modeling spatiotemporal dynamics in real-world videos. To address this limitation, in this study, we introduce Generalizable Implicit Motion Modeling (GIMM), a novel and effective approach to motion modeling for VFI. Specifically, to enable GIMM as an effective motion modeling paradigm, we design a motion encoding pipeline to model spatiotemporal motion latent from bidirectional flows extracted from pre-trained flow estimators, effectively representing input-specific motion priors. Then, we implicitly predict arbitrary-timestep optical flows within two adjacent input frames via an adaptive coordinate-based neural network, with spatiotemporal coordinates and motion latent as inputs. 
Our GIMM can be easily integrated with existing flow-based VFI works by supplying accurately modeled motion.
We show that GIMM performs better than the current state of the art on standard VFI benchmarks.
\end{abstract}

\section{Introduction}
\label{sec:intro}
Video Frame Interpolation (VFI) is a fundamental task in computer vision, which involves generating intermediate frames between two adjacent video frames. This technique is crucial for various practical applications, including novel view synthesis \cite{zhou2016view, flynn2016deepstereo, li2021nsff}, video generation \cite{singer2022make}, and video compression \cite{wu2018video}.
This task is highly challenging due to the complex motions typically found in real-world videos. To address this, recent research \cite{li2023amt, zhang2023emavfi, reda2022film, park2021abme, huang2022rife} has focused on flow-based frameworks, which have shown substantial success. Generally, these frameworks for VFI involve two main phases: 1) transforming the input frames based on estimated optical flows, and 2) merging and enhancing the warped frames to produce intermediate frames. Consequently, the accuracy of flow estimation is crucial for the fidelity of the synthesized frames.

\begin{figure*}[!t]
  \centering
  \includegraphics[width=1.0\linewidth]{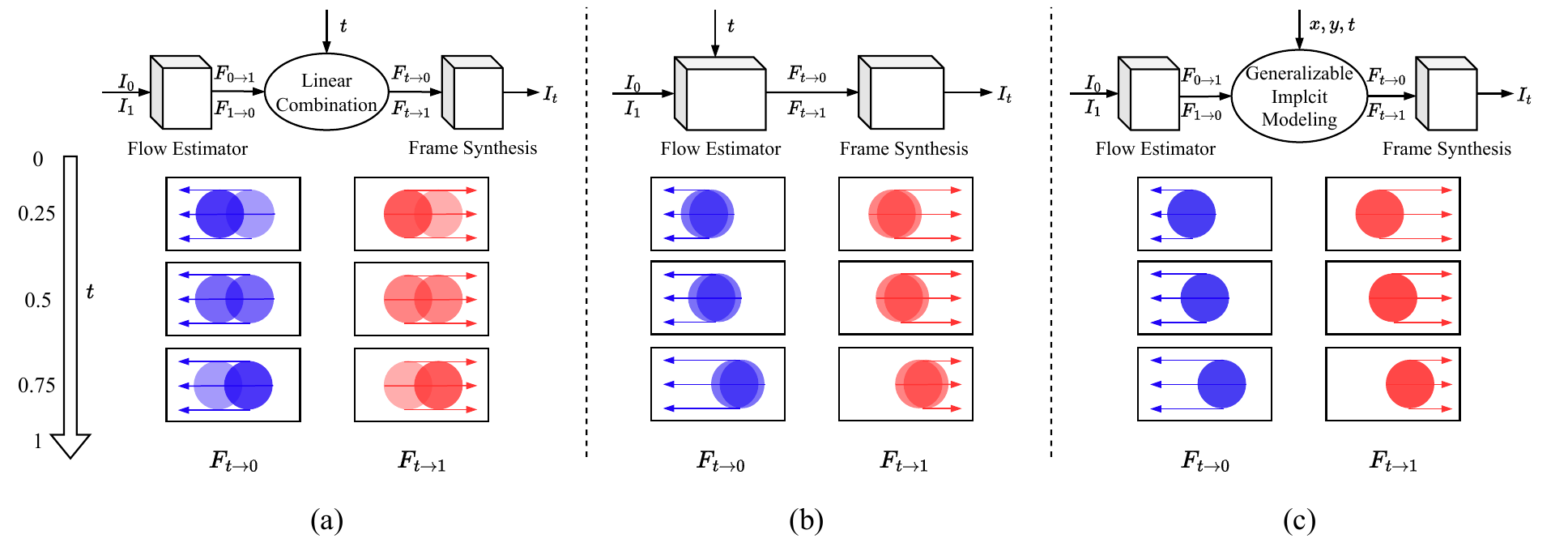}
  \caption{
  \textbf{Schematic of motion modeling paradigms} in video frame interpolation. (\textbf{a}) A na\"{i}ve \textit{linear combination} of bidirectional flows $F_{0\rightarrow1}, F_{1\rightarrow0}$ (i.e., flows between input frames) may lead to ambiguous and coarse motion estimation due to strong overlapped and linear assumptions. (\textbf{b}) A \textit{time-condition-based modeling} approach may predict suboptimal bilateral flows $F_{t\rightarrow0}, F_{t\rightarrow1}$ (i.e., flows between estimated and input frames), capturing spatiotemporal changes for moving objects ineffectively. (\textbf{c}) Our \textit{generalizable implicit motion modeling} properly represents spatiotemporal dynamics across videos and predict better bilateral flows via an adaptive coordinate-based neural network.
  }
  \label{fig:teaser}
\end{figure*}

Accurately modeling flow from two distant frames is challenging due to the complexities of capturing subtle and dynamic movements caused by varying motion speeds, object occlusions, and lighting conditions. A commonly used approach in the literature combines bidirectional flows derived from input frames \cite{jiang2018superslomo, park2021abme, shangguan2022cure, xu2019quadratic, BMBC} (Figure~\ref{fig:teaser}(a)). This method assumes \textit{overlapped} and linear motion for flow estimation, which does not accurately reflect real-world dynamics. Several studies \cite{huang2022rife, li2023amt, zhang2023emavfi, reda2022film} directly predict bilateral flows based on intermediate discrete timestamps (Figure~\ref{fig:teaser}(b)). These approaches model the correlation between frame motions and timestamps without leveraging motion priors within input frames; they thus fall short of capturing complex spatial-temporal changes and handling occluded regions and unexpected deformations. Moreover, this discrete-time-based modeling paradigm is ineffective for arbitrary-time interpolation.

In this study, we explore a more effective and generalizable implicit approach to motion modeling for VFI. Inspired by the success of implicit neural representations \cite{sitzmann2020siren} in encoding complex high-dimensional data such as 2D images \cite{liif2021}, 3D scenes \cite{mildenhall2020nerf}, and videos \cite{chen2021nerv}, we propose to implicitly model optical flows between two adjacent video frames using coordinate-based neural networks. This implicit paradigm can directly take arbitrary spatiotemporal coordinates as inputs and effectively decode the desired space-time outputs, making it a promising framework for learning highly dynamic optical flows in real-world videos. 
However, leveraging implicit neural networks for effective motion modeling poses challenges. First, standard implicit neural networks typically perform per-instance modeling, optimizing model parameters for a single specific input. This limitation restricts their applicability across different input video frames. Therefore, we need to develop a more adaptive implicit model capable of capturing motions in any given video. Second, efficiently integrating spatiotemporal information within implicit neural networks is complex, especially when dealing with the intricate motions occurring between two video frames. This necessitates designing appropriate implicit neural architectures that can accurately predict and represent both the spatial and temporal dynamics in videos.

To this end, we propose a novel generalizable implicit flow encoding for motion modeling in VFI, called Generalizable Implicit Motion Modeling (GIMM) (Figure~\ref{fig:teaser}(c)). Our method only assumes the availability of bidirectional flows ($F_{0\rightarrow1}, F_{1\rightarrow0}$ ) of two input frames obtained from a pre-trained optical flow estimator (e.g., RAFT \cite{teed2020raft}, FlowFormer \cite{huang2022flowformer}). The input flows can be noisy as this prior will be refined to estimate the bilateral flows ($F_{t\rightarrow0}, F_{t\rightarrow1}$) between arbitrary intermediate timestamps. Our method is unique in that it can represent input-specific motion priors effectively. For instance, it can accurately capture the complex dynamics of a somersault. This is achieved by introducing a motion encoding pipeline to extract spatiotemporal motion latent from the bidirectional flows. Our method can further enable frame interpolation at \textit{arbitrary timestamps}, thanks to the adaptive coordinate-based neural network that takes spatiotemporal coordinates and motion latent as inputs. This capability allows our method to generate frames at various temporal granularities, providing flexibility and precision in video frame interpolation.

Our contributions are summarized as follows: We present an effective motion modeling paradigm for video frame interpolation characterized by a novel generalizable implicit motion modeling framework. Our GIMM is capable of accurately predicting optical flow for arbitrary timesteps between two adjacent video frames at any resolution, allowing seamless integration with existing flow-based VFI methods. We demonstrate the advantages of our GIMM in motion modeling for arbitrary-timestep VFI tasks, achieving state-of-the-art performances on various benchmarks.

\section{Related Work}
\label{sec:relwork}
\noindent\textbf{Video frame interpolation.} 
Conventional VFI studies primarily rely on either direct frame synthesis via convolutional networks \cite{choi2020snufilm, DAIN, kalluri2023flavr} or interpolation using dynamic kernels with learnable weights and offsets \cite{Niklaus_CVPR_2017, Niklaus_ICCV_2017, peleg2019net, EDSC, cheng2020video, ding2021cdfi, lee2020adacof}. Recent approaches have shifted towards flow-based methods to synthesize frames at desired timesteps, where motion modeling plays a crucial role \cite{huang2022rife, li2023amt, xue2019vimeo90k, Kong_2022_ifrnet, siyao2021deep, hou2023video}. Some flow-based methods combine estimated bidirectional flows between input frames \cite{jiang2018superslomo, park2021abme, shangguan2022cure, xu2019quadratic, BMBC, DAIN}, often leading to inaccurate motion predictions, especially in occluded areas. This simplistic combination results in ambiguous and coarse motion estimation, causing object shifts in interpolated frames.
Several recent approaches \cite{huang2022rife, li2023amt, zhang2023emavfi, huang2022rife, lu2022vfiformer} address these issues by directly predicting the desired motion within an end-to-end framework conditioned on timesteps, showing impressive results in synthesized frames. However, these methods mainly rely on discrete-time-based fitting of variable relationships between motion and timesteps, making it challenging to achieve consistent, continuous interpolation outcomes.

\noindent\textbf{Implicit neural representations.}
Implicit Neural Representations (INRs) have been shown effective in modeling complex, high-dimensional data for various applications, including video compression \cite{chen2021nerv}, novel view synthesis \cite{mildenhall2020nerf, li2021nsff}, and image super-resolution \cite{liif2021}. Typically, INRs learn a continuous mapping from a set of coordinates to a specific signal using a coordinate-based neural network to encode data implicitly. The flexible and expressive modeling capabilities of INRs motivate us to explore their use to encode intricate motions and capture subtle, dynamic movements of objects in real-world videos.
A recent work related to ours is IFE \cite{figueiredo2023frameife}. While both approaches consider implicit flow encoding, our method significantly differs from IFE. Unlike IFE, which focuses on per-scene encoding—where each coordinate-based network is parameterized by a specific video instance—we aim to develop a generalizable motion modeling approach that can be applied across different videos.

\noindent\textbf{Generalizable INRs.} Recent works~\cite{liif2021,saito2019pifu,chibane2020implicit,jiang2020local,genova2020local,chen2022videoinr,chen2023motif} further extend INRs for generalizable encoding by conditioning coordinate-based neural networks with additional instance-specific inputs. For example, some approaches~\cite{chen2022transinr,lee2024locality,kim2023ginripc} employ Transformers~\cite{vaswani2017attention} as meta-learners to predict instance-specific weights or modulation features for coordinate-based neural networks at high computational costs. Several notable studies \cite{chen2023motif,chen2022videoinr,shangguan2022cure} leverage generalizable INRs for video encoding to facilitate video interpolation and super-resolution, mainly focusing on directly learning implicit space-time continuous neural representations from video. 
In contrast, we aim to explore effective motion modeling paradigms to improve intermediate frame synthesis for flow-based VFI. To our knowledge, we make the first attempt to utilize generalizable INRs for motion modeling in the context of VFI.

\section{Method}\label{sec:method}

Given a pair of adjacent video frames $I_0, I_1 \in \mathbb{R}^{H \times W \times3}$ with timesteps $\{0, 1\}$, a general flow-based video frame interpolation process is defined as follows:
\begin{align}
  & F_t = \mathcal{G}(I_0,I_1,t),\\
  & I_t = \mathcal{H}(F_t,I_0,I_1),\forall t \in [0,1],
\label{eq:general1}
\end{align}
where $\mathcal{G}$ denotes a motion modeling (or flow estimation) module, and $\mathcal{H}$ indicates a frame interpolation process under the guidance of estimated motion $F_t$ at a timestep of $t \in [0,1]$ for input frames $I_0$ and $I_1$. In this work, we mainly focus on studying an effective motion modeling framework $\mathcal{G}$ for flow-based VFI.

\subsection{Generalizable Implicit Motion Modeling}
\label{sec:gimm}

\begin{figure*}[h!]
  \centering
  \includegraphics[width=1.0\linewidth]{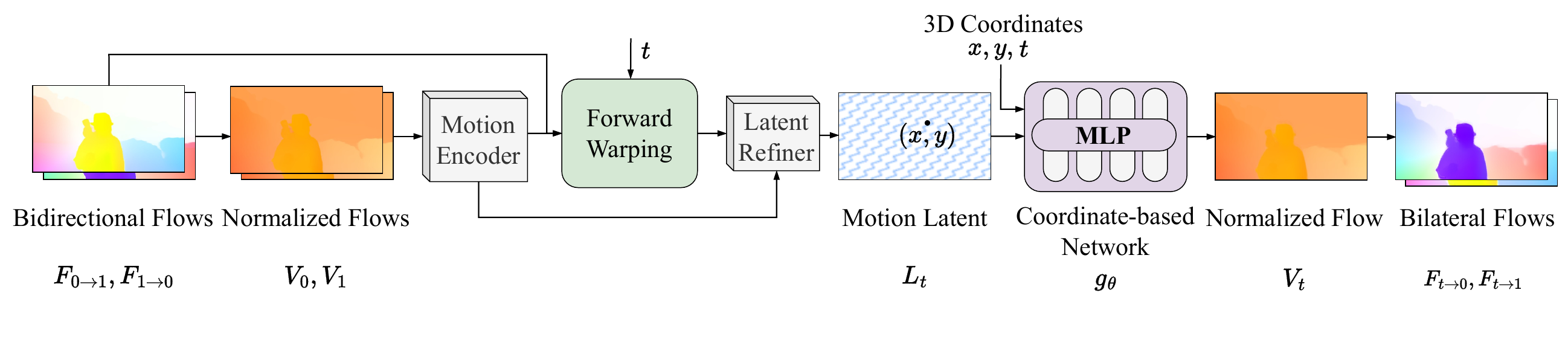}
  \caption{Our GIMM first transforms initial bidirectional flows $F_{0\rightarrow1},F_{1\rightarrow0}$ as normalized flows $V_0, V_1$. The motion encoder extracts motion features $K_0, K_1$ from $V_0, V_1$, respectively. $K_0, K_1$ are then forward warped at a given timestep $t$ using bidirectional flows to obtain the warped features $K_{t\rightarrow0}, K_{t\rightarrow1}$. We pass the warped and initial motion features into a Latent Refiner that outputs motion latent $L_t$, representing motion information at $t$. Conditioned on $L_t(x,y)$, the coordinate-based network $g_{\theta}$ predicts the corresponding normalized flow $V_t$ with 3D coordinates $\textbf{x}=(x,y,t)$. For interpolation usage, $V_t$ is then transferred into bilateral flows $F_{t\rightarrow0},F_{t\rightarrow1}$ through denormalization.}
  \label{fig:gimm}
\end{figure*}

The goal of GIMM is to estimate bilateral flows $F_{t\rightarrow0}, F_{t\rightarrow1}$ for any timestep $t\in[0,1]$, deriving from initial bidirectional flows $F_{0\rightarrow1}, F_{1\rightarrow0}$:
\begin{equation}
  F_{t\rightarrow0}, F_{t\rightarrow1} = \mathcal{G}(F_{0\rightarrow1}, F_{1\rightarrow0}, t), \forall t \in [0,1], 
  \label{eq:gimm}
\end{equation}
where $F_{0\rightarrow1}, F_{1\rightarrow0}$ are predicted from a pre-trained optical flow estimator (e.g., RAFT~\cite{teed2020raft}, FlowFormer~\cite{huang2022flowformer}) with input frames $I_0$ and $I_1$. Figure \ref{fig:gimm} depicts the overall generalizable motion modeling framework of GIMM. Motivated by the great success of INRs in modeling complex video data~\cite{chen2022videoinr, chen2023motif, figueiredo2023frameife, mildenhall2020nerf}, GIMM uses an adaptive coordinate-based neural network for continuous motion modeling. Unlike the existing IFE~\cite{figueiredo2023frameife} that performs per-scene optimization and requires test-time learning for different videos, our GIMM takes additional instance-specific motion latent inputs $L_t$ to enhance model generalizability across various input videos.

\noindent\textbf{Flow normalization.}
\label{sec:nm}
Following IFE~\cite{figueiredo2023frameife}, we perform normalization for the initial bidirectional flows as follows:
\begin{align}
    & V_0 = \phi(F_{0\rightarrow1}), V_1 = \phi(-F_{1\rightarrow0}),\\
    & V_t = \phi(F_{t\rightarrow1}-F_{t\rightarrow0}),
\end{align}
where $\phi$ is a scale operator. 
This reversible normalization process aligns the scale and temporal direction of input bidirectional flows $F_{0\rightarrow1}, F_{1\rightarrow0}$ with output bilateral flows $F_{t\rightarrow0}, F_{t\rightarrow1}$, allowing the normalized flows $V_i \in [0,1]^{H\times W \times 2}, i\in {0, 1}$ to be effectively encoded in subsequent implicit motion modeling in GIMM.

\noindent\textbf{Motion latent.}
\label{sec:motionlatent}
To achieve generalizable implicit modeling of motion dynamics, we learn implicit neural representations for motions with conditions on instance-specific motion latent inputs $L_t(x,y)$, which provides instance-specific motion priors at spatiotemporal coordinates $\textbf{x}=(x,y,t)$ for target motion $V_t(x,y)$. 
Specifically, we introduce a Motion Encoder to extract motion features ${K}_i$ from normalized flows $V_i$. 
To maintain spatiotemporal consistencies of modeled motion, we derive time-dependent motion features $K_t$ for given timestep $t$ from both motion features ${K}_0$ and ${K}_1$. To realize this, we apply forward warping $\overrightarrow{\omega}$ process to map every location of motion features ${K}_i$ to the target timestep $t$:
\begin{align}
    & K_{i\rightarrow t} = \overrightarrow{\omega}(K_i,F_{i\rightarrow t},Z_i).
    \label{eq:fwarp}
\end{align}
Here, $K_{i\rightarrow t}$ are warped features with timesteps $i\in \{0, 1\}$ that integrate the correspondences from source (input) timesteps to the target timestep, $Z_i$ represents splatting weights, and $F_{i\rightarrow t}$ denotes scaled bidirectional flows at timestep $t$. We compute scaled bidirectional flows $F_{i\rightarrow t}$ as follows:
\begin{align}
F_{i\rightarrow t} =
\begin{cases}
(t-0) \cdot F_{0\rightarrow 1},  & \text{if}\ i=0 \\
(1-t) \cdot F_{1\rightarrow 0}, & \text{if}\ i=1
\end{cases}
\label{eq:scale}
\end{align}
To mitigate warping errors in multi-to-one cases or occluded regions, we calculate splatting weights $Z_i$ using flow consistency $U^i_{flow}$ and variance $U^i_{var}$ metrics~\cite{niklaus2023splattingwacv,niklaus2020softmax} as follows:
\begin{align}
  & Z_i = \frac{1}{1+\alpha_{flow} \cdot U^i_{flow}} + \frac{1}{1+\alpha_{var} \cdot U^i_{var}},
  \label{eq:weights}
\end{align}
where $\alpha_{flow}$ and $\alpha_{var}$ are learnable parameters.
To obtain the final motion latent $L_t$, we concatenate the warped features $\texttt{Concat}(K_{0\rightarrow t}, K_{1\rightarrow t})$ as the coarse motion latent at timestep $t$ and further refine via a Latent Refiner module to deal with potential information loss and ambiguous motion in the forward warping process. Both the Latent Refiner and the Motion Encoder mentioned above are structured as shallow convolutional networks. We provide their network configurations and detailed calculations for two flow metrics $U^i_{flow}$ and $U^i_{var}$ in the supplementary.

\noindent\textbf{Implicit motion prediction.}
To realize INRs for generalizable motion modeling, we devise an adaptive coordinate-based network $g_{\theta}$ for implicit and continuous motion encoding, which maps spatiotemporal coordinates $\textbf{x} \in [-1,1]^{H\times W\times3}$ along with the corresponding $D$-dimension motion latent code $L_t \in \mathbb{R}^{H \times W \times D} $ to predicted normalized flows $\hat{V_t}$:
\begin{align}
  & \hat{V_t} = g_{\theta}(\textbf{x},L_t).
\label{eq:imp}
\end{align}
The predicted normalized flow $V_t$ can be converted into predicted bilateral flows $F_{t\rightarrow0}, F_{t\rightarrow1}$ via the reverse flow normalization process for VFI. See the supplementary for the details of the coordinate-based network.

\noindent\textbf{Optimization.}
In practice, we optimize the GIMM module by minimizing a Mean-Square-Error (MSE) loss between predicted ($\hat{V_t}$) and ground-truth ($V_t$) normalized optical flows:
\begin{align}
& \mathcal{L}_{gimm}(V_t,\hat{V_t}) = \frac{1}{H\times W}\sum^{H-1}_{k=0} \sum^{W-1}_{l=0} ||V_t(k,l)-\hat{V_t}(k,l)||_2.
\label{eq:gimm_mse}
\end{align}

\begin{figure*}[!t]
  \centering
  \includegraphics[width=1.0\linewidth]{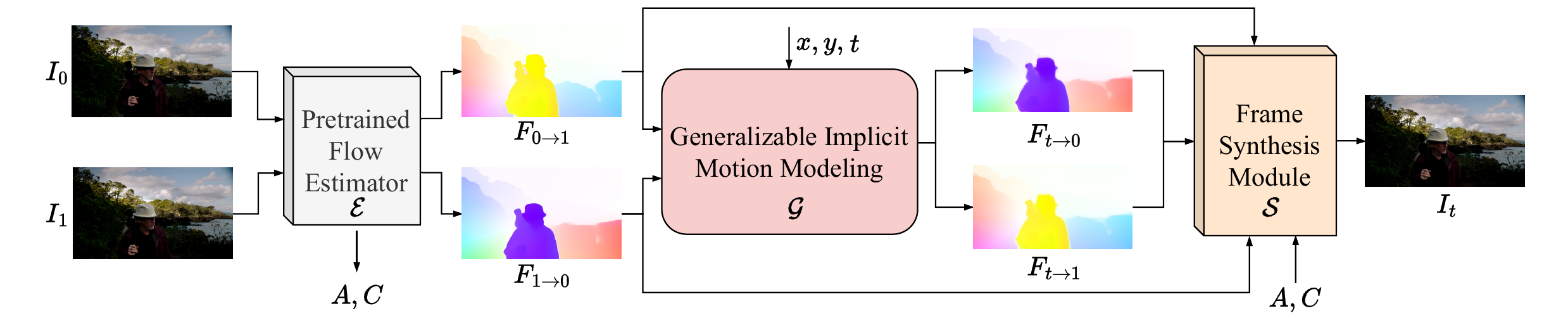}
  \caption{An overview of GIMM-VFI architecture. GIMM-VFI employs a pre-trained flow estimator, $\mathcal{E}$, to predict bidirectional flows $(F_{0\rightarrow1}, F_{1\rightarrow0})$ and extracts context features $A$ as well as correlation features $C$ from the input frames $(I_0, I_1)$. Given the timestep $t$, a generalizable implicit motion modeling (GIMM) module $\mathcal{G}$ (detailed in Figure \ref{fig:gimm}) takes the bidirectional flows as inputs and predicts bilateral flows $(F_{t\rightarrow0}, F_{t\rightarrow1})$, which are then passed into a frame synthesis module $\mathcal{S}$, together with extracted features $(A, C)$, to synthesize the target frame $I_t$.}
  \label{fig:main}
\end{figure*}
\subsection{Integrating GIMM with Frame Interpolation}
\label{sec:fip}
\noindent\textbf{GIMM-VFI.} 
Given two adjacent video frames $I_i$ with $i\in \{0, 1\}$,
we start with a pre-trained optical flow estimator (e.g., RAFT~\cite{teed2020raft}, FlowFormer~\cite{huang2022flowformer}) to extract context features $A_i$, correlation features $C_i$, and initial bidirectional flows $F_{0\rightarrow1}$ and $F_{1\rightarrow0}$. We first predict bilateral flows $F_{t\rightarrow i}$ for a given timestep $t$ from initial bidirectional flows using a pre-trained GIMM module. 
The predicted bilateral flows $F_{t\rightarrow i}$, context features $A_i$, and correlation features $C_i$ are then passed into a frame synthesis module $\mathcal{S}$ (adapted from the AMT~\cite{li2023amt}), which refines the input flows and generates warping masks $M_t\in [0,1]^{H\times W}$.
Following previous flow-based VFI studies~\cite{li2023amt,zhang2023emavfi,reda2022film,huang2022rife}, we obtain warped images $I_{t\rightarrow i}$ via backward warping $\overleftarrow{\omega}$ under the guidance of predicted bilateral flows  $F_{t\rightarrow i}$ from input frames $I_0$ and $I_1$: 
\begin{align}
& I_{t\rightarrow i} = \overleftarrow{\omega}(I_i,F_{t\rightarrow i}),\\
& I_{t\rightarrow i}(x,y) = I_i(x+F_{t\rightarrow i}^h(x,y),y+F_{t\rightarrow i}^v(x,y)), 
\label{eq:bwarp_frame}
\end{align}
where $F_{t\rightarrow i}^h,F_{t\rightarrow i}^v$ represents horizontal and vertical motion of $F_{t\rightarrow i}$, respectively.
To generate the final interpolated image $\hat{I_t}$, the warped images $I_{t\rightarrow 0},I_{t\rightarrow 1}$ are fused with the warping mask $M_t$ as follows:
\begin{align}
& \hat{I_t} = M_t\cdot I_{t\rightarrow0} + (1-M_t) \cdot I_{t\rightarrow1}.
\label{eq:final_interp}
\end{align}
See the supplementary for the details of the frame synthesis module.

\noindent\textbf{Overall objectives.} 
We use the following objective function for VFI: 
\begin{equation}
\mathcal{L}_{interp}(I_t,\hat{I_t}) = \mathcal{L}_{lap}(I_t,\hat{I_t})+\mathcal{L}_{char}(I_t,\hat{I_t})+\mathcal{L}_{census}(I_t,\hat{I_t}),
\label{eq:loss_inter}
\end{equation}
where $\mathcal{L}_{lap}$, $\mathcal{L}_{char}$, and $\mathcal{L}_{census}$ denote the Laplacian loss~\cite{zhang2023emavfi,niklaus2020softmax}, Charbonnier loss~\cite{li2023amt,charbonnier1994two} , and census loss~\cite{li2023amt,meister2018unflow}, respectively.
In addition, to preserve the motion modeling in the pre-trained GIMM module, we reconstruct and supervise the flows $\hat{V_0},\hat{V_1}$ with pseudo ground truth $V_0, V_1$.
This objective $\mathcal{L}_{rec}$ is defined  as $||V_0-\hat{V_0}||_2 + ||V_1-\hat{V_1}||_2$.
The overall objective $\mathcal{L}$ is as follows:
\begin{align}
& \mathcal{L} = \mathcal{L}_{interp}(I_t,\hat{I_t})  + \lambda_{rec}\mathcal{L}_{rec},
\label{eq:loss_all}
\end{align}
where $\lambda_{rec}$ is the hyperparameter. We optimize the entire GIMM-VFI model that contains the pre-trained flow estimator, pre-trained GIMM module, and frame synthesis module.

\section{Experiments}\label{sec:exp}
We present quantitative and qualitative evaluations of our motion modeling method GIMM in Section~\ref{sec:motion_exp}, and the corresponding interpolation method (GIMM-VFI) in Section~\ref{sec:interpo}. Specifically, we evaluate both motion quality and performance on the downstream interpolation task. We compare GIMM-VFI with current state-of-the-art VFI methods on arbitrary-timestep interpolation.

\noindent\textbf{Implementation details.} We train the GIMM model on the training split of Vimeo90K~\cite{xue2019vimeo90k} triplets dataset using optical flows extracted by off-the-shelf flow estimators. Our GIMM-VFI is trained on the complete Vimeo90K septuplet dataset. Specifically, we implement two variants of GIMM-VFI, using two different flow estimators: the RAFT~\cite{teed2020raft} and FlowFormer~\cite{huang2022flowformer}, designated as GIMM-VFI-R and GIMM-VFI-F, respectively. More details are provided in the supplementary materials.

\subsection{Motion Modeling}\label{sec:motion_exp}
\noindent\textbf{VTF and VSF benchmarks.} To assess the modeled motion quality with GIMM, we use Flowformer~\cite{huang2022flowformer} to produce pseudo ground truth motion. Specifically, we establish two motion evaluation benchmarks: Vimeo-Triplet-Flow (VTF) and Vimeo-Septuplet-Flow (VSF). These benchmarks are derived from the triplet and septuplet splits of the Vimeo90K~\cite{xue2019vimeo90k} test set, tailored for evaluating 2X and 6X motion modeling, respectively. 

\noindent\textbf{Baselines.}
We compare GIMM with other motion modeling approaches, including \textit{Linear}, \textit{Forward Warp}, \textit{End-to-End}, and BMBC~\cite{BMBC}.
Specifically, \textit{Linear} replaces our GIMM module with the linear approximation strategy~\cite{jiang2018superslomo}. Similarly, \textit{Forward Warp} substitutes GIMM with a forward warping strategy~\cite{niklaus2023splattingwacv}. Meanwhile, the \textit{End-to-End} denotes the strategy that directly predicts motion at arbitrary timesteps through an end-to-end fitting.
In this end-to-end setting, we select the current state-of-the-art EMA-VFI~\cite{zhang2023emavfi} as the representative method. 
Additionally, we make further comparisons with BMBC~\citep{BMBC}, which is supposed to model complex motion effectively through several stages of motion predictions and motion refinements.
For fair comparisons,  we employ RAFT~\cite{teed2020raft} as the flow estimator for different motion modeling methods.

\noindent\textbf{Evaluation settings.}
We measure the modeled motion quality on both VTF and VSF benchmarks by calculating PSNR on normalized flows and End-Point-Error (EPE) on the unscaled flows. For the downstream interpolation task, we calculate PSNR on the `hard' split of the SNU-FILM-arb benchmark (SNU-FILM-arb-Hard). Details about SNU-FILM-arb are presented in Section~\ref{sec:interpo}.

\begin{table}[!t]
    \caption{\textbf{Comparisons of different motion modeling methods.} We assess the modeled motion on Vimeo-Triplet-Flow (VTF) and Vimeo-Septuplet-Flow (VSF) by employing PSNR and EPE metrics. Additionally, we demonstrate their impact on the interpolation task by presenting the PSNR values of their interpolation results on the `hard' split of the SNU-FILM-arb dataset. We denote the calculated PSNR values as PSNR\textsubscript{f} for optical flows in the motion assessment and PSNR\textsubscript{i} for interpolated images in the interpolation task.
    }
    \small
    \centering
    \begin{tabular}{lccccc}
    \toprule
    \multirow{2}{*}{Motion method} & \multicolumn{2}{c}{Vimeo-Triplet-Flow (VTF)} & \multicolumn{2}{c}{Vimeo-Septuplet-Flow (VSF)} &SNU-FILM-arb-Hard\\
    \cmidrule(lr){2-6}
     & PSNR\textsubscript{f}$\uparrow$ & EPE$\downarrow$ & PSNR\textsubscript{f}$\uparrow$ & EPE$\downarrow$ & PSNR\textsubscript{i}$\uparrow$ \\
    \midrule
    Linear & 35.03 & 0.44 & 30.09 & 2.87 & 32.42 \\
    Forward Warp  & 32.80 & 0.47 & 28.22 & 3.38 & 32.31 \\
    End-to-End & 29.23 & 1.02& 25.99 & 5.12  & 32.28\\
    BMBC~\citep{BMBC} & 28.89 & 0.95 & 23.19 & 8.23 & 28.51 \\
    GIMM (-VFI-R) & \textbf{37.56} & \textbf{0.34} & \textbf{30.45} & \textbf{2.68} & \textbf{32.62}\\
    \bottomrule
    \end{tabular}
    \label{tab:motion_modeling_comparison}
\end{table}
\begin{figure*}[!t]
  \centering
  \includegraphics[width=1.0\linewidth]{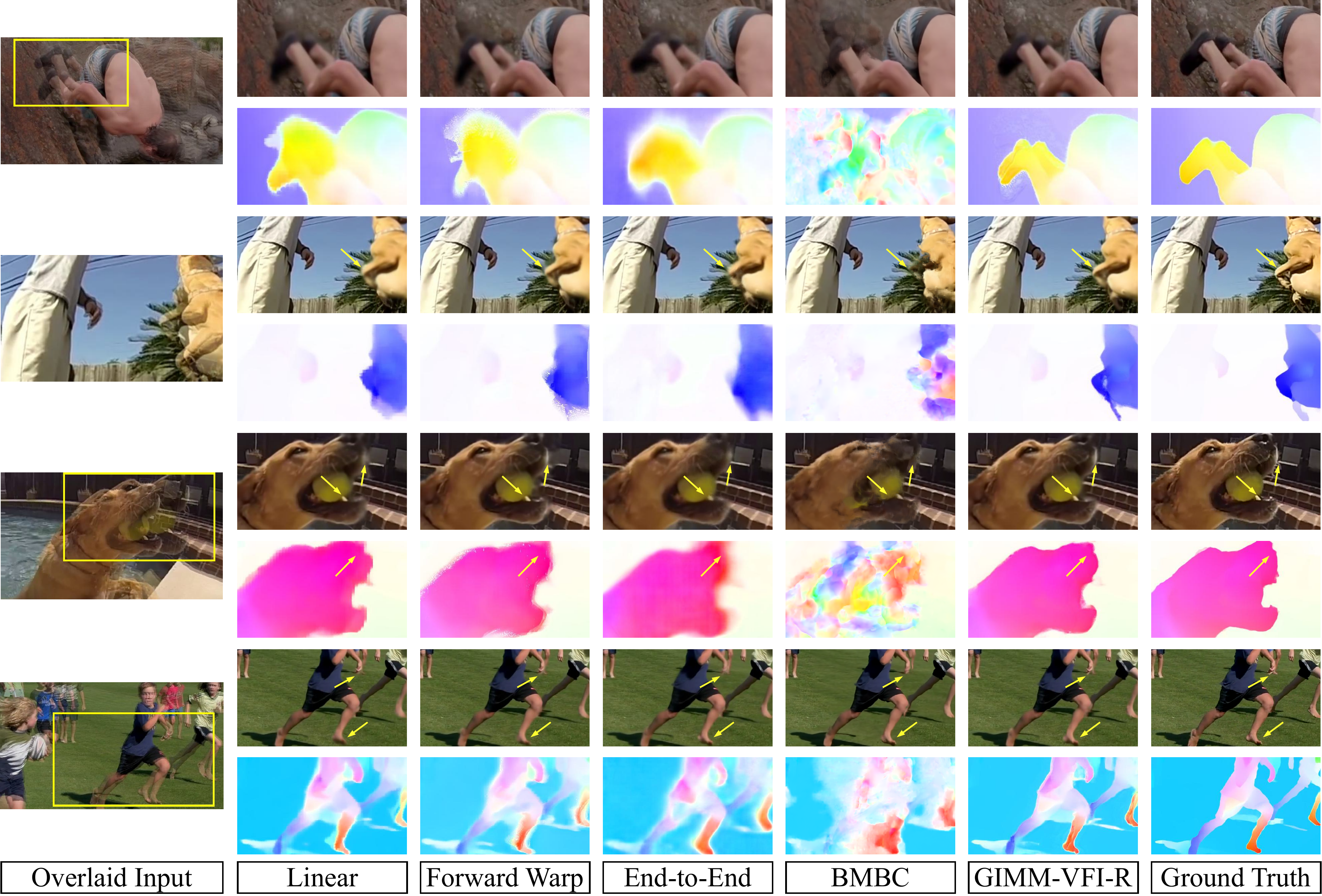}
  \caption{Qualitative comparisons of different motion modeling methods on SNU-FILM-arb-Hard. All the results are predicted at $t=0.75$, and ground truth flows are obtained by FlowFormer~\cite{huang2022flowformer}. }
  \label{fig:motion_flow_interp}
\end{figure*}

\noindent\textbf{Results.} We provide quantitative comparisons of our GIMM and baselines in Table~\ref{tab:motion_modeling_comparison}. We first report the results of motion quality on VTF and VSF. We find that GIMM can continuously model motions in videos. GIMM predicts the best flows on both the VTF benchmark (37.56dB PSNR/ 0.34 EPE) and the VSF benchmark (30.45dB PSNR/ 2.68 EPE) that involves unseen timesteps during training. Moreover, our study highlights the importance of motion priors in both motion modeling and the downstream interpolation task. The \textit{Linear}, \textit{Forward Warp}, and our GIMM methods that leverage motion priors from RAFT~\cite{teed2020raft} can provide better motion and interpolation performances than the \textit{End-to-End} and BMBC~\cite{BMBC}
across all benchmarks. Notably, GIMM benefits the interpolation task the most and achieves the highest PSNR of 32.62 dB on SNU-FILM-arb-Hard. While motion modeling plays a critical role in interpolation, other modules also impact interpolation outcomes. The \textit{End-to-End} method has a narrower performance gap with other advanced motion modeling approaches on the interpolation benchmark than on motion modeling benchmarks, likely due to its more complex and extensive feature extraction and frame synthesis modules.

\noindent\textbf{Visualizations.} Besides the quantitative evaluation of motion modeling, we also qualitatively evaluate them on SNU-FILM-arb-Hard, as shown in Figure~\ref{fig:motion_flow_interp}. Since SNU-FILM-arb-Hard requires 8X interpolation for input frames of large motion, classical methods like BMBC~\cite{BMBC} may struggle to model the intermediate motion. In contrast, we observe the capacity of GIMM for accurate motion modeling at arbitrary timesteps. Take the Somersault in Figure~\ref{fig:motion_flow_interp} for example, the ankle flow and the corresponding interpolation generated by GIMM have the best consistency with the ground truth, pointing to the same direction. Moreover, GIMM is capable of reducing ambiguities when interpolating moving objects. We can observe clearer silhouettes with fewer blurs in both the `dog-leg' and `dog-head' cases in Figure~\ref{fig:motion_flow_interp}.

\begin{table}[!t]
    \caption{\textbf{Quantitative comparisons of different motion model choices.}
    }
    \small
    \centering
    \begin{tabular}{lccccc}
    \toprule
    \multirow{2}{*}{Model Variants} & \multicolumn{2}{c}{Vimeo-Triplet-Flow (VTF)} & \multicolumn{2}{c}{Vimeo-Septuplet-Flow (VSF)} & \multirow{2}{*}{Param.(M)}\\
    \cmidrule(lr){2-5}
     & PSNR$\uparrow$ & EPE$\downarrow$ & PSNR$\uparrow$ & EPE$\downarrow$ & \\
    \midrule
    Meta-learning (INR) & 30.19 & 0.88 & 24.50 & 6.80 & 43.92\\
    GIMM (U-Net) & 36.96 & 0.39  & 29.96 & 2.90 & 4.27\\
    GIMM (INR) & \textbf{37.56} & \textbf{0.34} & \textbf{30.45} & \textbf{2.68} & \textbf{0.25}\\
    \bottomrule
    \end{tabular}
    \label{tab:sup_ablation}
\end{table}

\subsection{Ablation Study on Motion Model Choices}
\label{sec:motion_model_ablation}
As our core contribution, we present an effective motion modeling paradigm
for video frame interpolation, characterized by a novel generalizable implicit motion modeling framework (GIMM). In this section, we further investigate the necessity of implicit neural representations (INRs) and approaches to generalizable INRs. The experiments are conducted on the motion modeling benchmarks, \textit{i.e.,} Vimeo-Triplet-Flow (VTF) and Vimeo-Septuplet-Flow (VSF).

\noindent\textbf{Necessity of INRs.}
A straightforward replacement of INRs in our GIMM is a timestep-conditioned U-Net as in diffusion literature~\cite{rombach2022ldm}. We substitute the INRs with a U-Net of a comparable number of channels to ensure fair comparisons.
In Table~\ref{tab:sup_ablation}, replacing INR with U-Net results in worse performance, especially on the 6X motion modeling benchmark VSF. This demonstrates the strong continuous modeling ability of INR. Besides, GIMM with INR has a much lighter architecture, improving efficiency. Therefore, it is necessary and proper to use INR for our motion modeling.

\noindent\textbf{Approaches to generalizable INRs.}
To realize generalizable implicit modeling, an alternative way is to modulate the weights of INRs with meta-learners~\cite{chen2022transinr,kim2023ginripc}. Following ~\cite{kim2023ginripc}, we employ Transformers~\cite{vaswani2017attention} as the meta-learner. Compared to GIMM, this meta-learning approach performs significantly worse across all benchmarks while introducing a model with over 170X more parameters.

\subsection{Ablation Study on GIMM}
In this section, we investigate the effectiveness of model designs in GIMM in Table~\ref{tab:ablation}. The experiments are conducted on the VTF and VSF benchmarks.

\noindent\textbf{Forward warping.} 
We substitute forward warping operation in GIMM with a straightforward linear combination of the input motion features (\textit{Non-Fwarp}). In Table~\ref{tab:ablation}, we observe a significant performance drop with this modification.  For example, the EPE on VSF increases by 0.20. This result highlights the importance of forward warping for GIMM's continuous motion modeling.

\noindent\textbf{Implicit modeling.} In GIMM, the motion can be modeled even without the presence of implicit modeling. We conduct experiments without using any coordinates (\textit{Non-Imp}). In Table~\ref{tab:ablation}, this variant yields performance gains, with reductions of 0.05 and 0.13 in the EPEs on VTF and VSF respectively, highlighting the importance of implicit modeling in GIMM.

\noindent\textbf{Motion encoder.} We leverage the motion encoder in our GIMM to extract features from flows, alleviating the negative impacts brought by the possible bias and noises in the estimated flows. To justify this design, we remove the motion encoder for a direct comparison (\textit{Non-ME}). As a result, we observe increases in error on both benchmarks, \textit{e.g.}, the EPE on VSF rises 0.17. Therefore, the motion encoder benefits the motion modeling in GIMM.

\noindent\textbf{Latent refinement.} 
To validate the efficacy of latent refinement in GIMM, we conducted experiments excluding the latent refiner (\textit{Non-Refiner}). This leads to notable declines of 0.53 dB and 0.43 dB in PSNRs across both benchmarks. Thus, refining the motion latent proves essential for accurate motion modeling.

\noindent\textbf{Spatial coordinates.} Since the modeled motion consists of spatiotemporal changes, we use 3D spatiotemporal coordinates in the implicit modeling of GIMM. To assess the necessity of spatial coordinates, we replace 3D coordinates with temporal coordinates (\textit{T-coord only}).
This removal causes a 0.16 dB drop of PSNR on Vimeo-Triplet-Flow and a 0.06 increase of EPE on Vimeo-Septuplet-Flow. This emphasizes the crucial role of spatial coordinates in implicit modeling.

\begin{table}[!t]
    \caption{\textbf{Quantitative comparisons of different model variants.} For each model variant, we evaluate its motion modeling performance on Vimeo-Triplet-Flow and Vimeo-Septuplet-Flow, respectively. We adopt EPE and PSNR as the metrics for motion quality.
    }
    \small
    \centering
    \begin{tabular}{lcccc}
    \toprule
    \multirow{2}{*}{Model Variants} & \multicolumn{2}{c}{Vimeo-Triplet-Flow (VTF)} & \multicolumn{2}{c}{Vimeo-Septuplet-Flow (VSF)} \\
    \cmidrule(lr){2-5}
     & PSNR$\uparrow$ & EPE$\downarrow$ & PSNR$\uparrow$ & EPE$\downarrow$ \\
    \midrule
    Non-Fwarp & 37.07 & 0.37 & 30.09 & 2.88\\
    Non-Imp & 37.04 & 0.39 & 30.11 & 2.81  \\
    Non-ME & 37.05 & 0.42 & 30.26 & 2.85\\
    Non-Refiner & 37.03 & 0.37 & 30.02 & 2.77\\
    T-coord only & 37.40 & 0.36  & 30.39 & 2.74\\
    Full & \textbf{37.56} & \textbf{0.34} & \textbf{30.45} & \textbf{2.68}\\
    \bottomrule
    \end{tabular}
    \label{tab:ablation}
\end{table}

\begin{table*}[!t]
    \caption{\textbf{Quantitative results for arbitrary-timestep interpolation.} We report the quantitative metrics reported as PSNR$\uparrow$/LPIPS$\downarrow$/FID$\downarrow$, with the best results highlighted in \sota{boldface} and the second best results in \subsota{underline}.
    The analysis categorizes methods into two groups: non-INR methods (first half) and INR-based methods (second half). We employ a `-' symbol to denote `out-of-memory' scenarios, and a dagger (`${\dag}$') to indicate that the results are drawn from prior studies~\cite{huang2022rife,hu2022m2m,zhang2023emavfi}.
    }
    \vspace{5pt}
    \centering
    \resizebox{\textwidth}{!}{
    \begin{tabular}{lccccc}
\toprule
         \multirow{2}{*}{Method} & \multicolumn{2}{c}{XTest}& \multicolumn{3}{c}{SNU-FILM-arb} \\
         \cmidrule(lr){2-3} \cmidrule(lr){4-6} 
         &2K & 4K & Medium (4X) & Hard (8X) & Extreme (16X) \\
\midrule
        RIFE~\cite{huang2022rife} & 31.43{\dag}~/~0.126~/~11.99 & 30.58{\dag}~/~0.152~/~13.52 & 36.33~/~0.038~/~6.65 & 31.87~/~0.072~/~11.99 & 27.21~/~0.134~/~19.82  \\ 
        M2M~\cite{hu2022m2m} & 32.13{\dag}~/~\subsota{0.098}~/~9.25 & 30.88{\dag}~/~0.158~/~8.67 & 36.56~/~0.036~/~5.98 & 31.92~/~0.061~/~10.13 & 27.14~/~0.112~/~17.37  \\ 
        IFRNet~\cite{Kong_2022_ifrnet} & 31.53{\dag}~/~0.108~/~23.93 & 30.46{\dag}~/~0.164~/~23.75 & 34.88~/~0.046~/~9.92 & 31.15~/~0.066~/~11.65 & 26.32~/~0.115~/~16.91\\ 
        AMT~\cite{li2023amt}  & 28.88~/~0.153~/~13.92 & 28.17~/~0.187~/~13.97 & 34.49~/~0.072~/~9.25 & 31.03~/~0.089~/~10.34 & 26.44~/~0.136~/~14.72 \\
        UPR-Net~\cite{jin2023uprnet}  & 31.16~/~0.104~/~10.75 & 30.50~/~0.154~/~9.45 & 36.78~/~\subsota{0.033}~/~6.09 & 31.96~/~0.064~/~\subsota{9.93} & 27.14~/~0.111~/~\subsota{16.76}  \\
        EMA-VFI~\cite{zhang2023emavfi} & 32.53~/~\sota{0.097}~/~7.21 & 31.21~/~0.156~/~8.61 & 36.65~/~0.041~/~7.07 & 32.28~/~0.074~/~12.17 & 27.72~/~0.130~/~19.58 \\
\midrule
        CURE~\cite{shangguan2022cure} & 30.24~/~0.111~/~26.42 & - & 36.09~/~0.035~/~6.98 & 31.32~/~0.063~/~12.72 & 26.61~/~0.114~/~22.62 \\ 
        GIMM-VFI-R &\subsota{32.71}~/~0.113~/~\sota{6.52} & \subsota{31.88}~/~\subsota{0.149}~/~\sota{6.49}	 & \subsota{37.02}~/~\subsota{0.033}~/~\subsota{5.89} & \sota{32.62}~/~\subsota{0.060}~/~\sota{9.59}	 & \subsota{27.99}~/~\subsota{0.110}~/~\sota{16.45}\\
        GIMM-VFI-F &\sota{32.91}~/~0.103~/~\subsota{6.74} & \sota{31.97}~/~\sota{0.142}~/~\subsota{6.58} & \sota{37.03}~/~\sota{0.031}~/~\sota{5.86}	 & \subsota{32.56}~/~\sota{0.059}~/~9.95	 & \sota{28.01}~/~\sota{0.109}~/~16.79\\
\bottomrule
    \end{tabular}
    }
    \label{tab:arbres}
\end{table*}

\begin{figure*}[!t]
  \centering
  \includegraphics[width=1.0\linewidth]{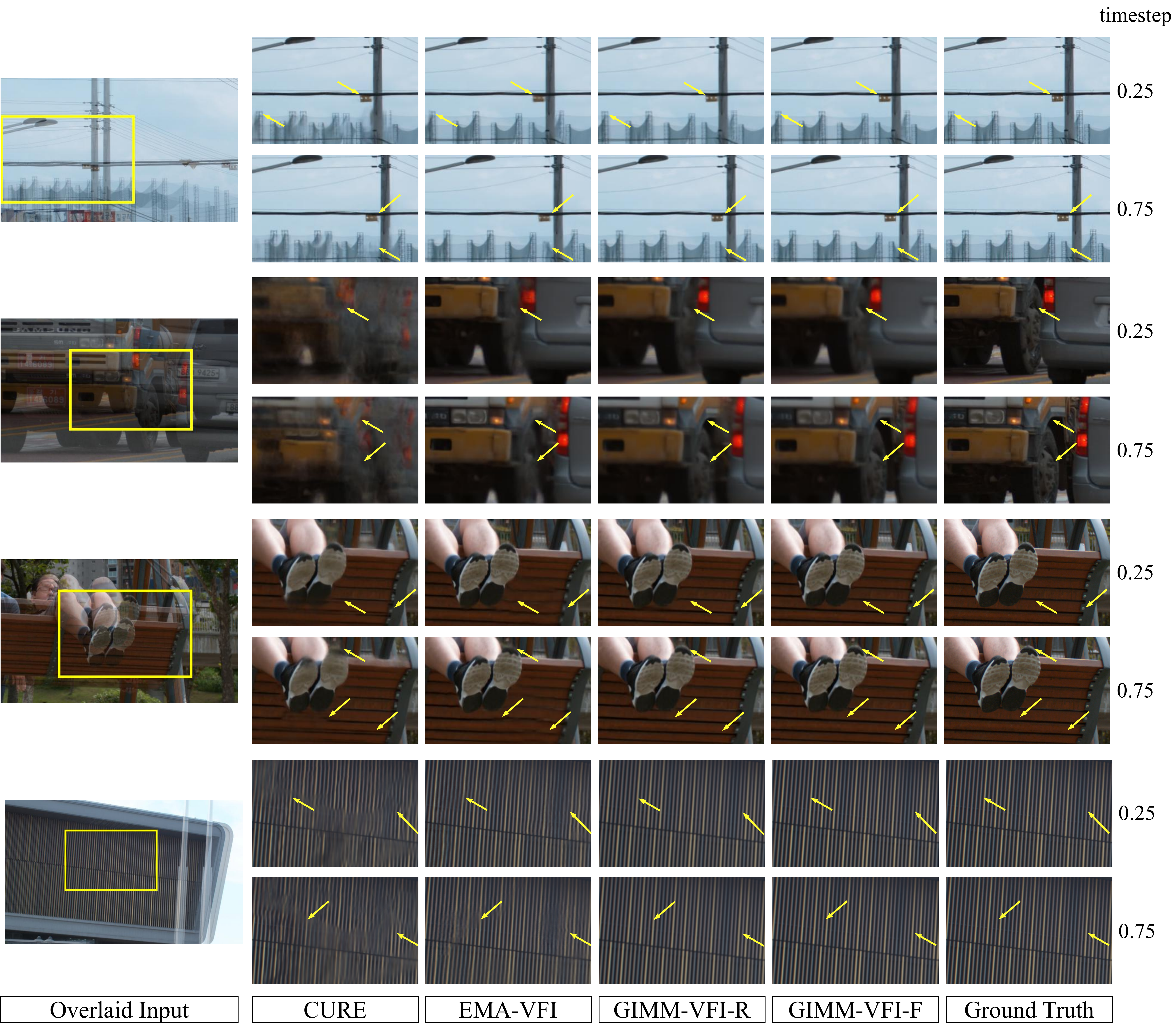}
  \caption{Qualitative comparisons of arbitrary-timestep interpolation on XTest-2K~\cite{sim2021xvfi}. Positions pointed by the yellow arrow indicate the distinct performance of our method.}
  \label{fig:visnx}
\end{figure*}

\subsection{Arbitrary-timestep Video Frame Interpolation}
\label{sec:interpo}
\noindent\textbf{SNU-FILM-arb benchmark.} We introduce the SNU-FILM-arb benchmark for a more generalized evaluation, to facilitate research on the arbitrary-timestep frame interpolation task. Specifically, we incorporate the SNU-FILM dataset~\cite{choi2020snufilm} that encompasses 310 clips from 31 videos captured at 240 fps, with heights ranging from 384 to 1280 pixels and widths from 368 to 720 pixels. Different from its original setting of 2X interpolation, we conduct 4X, 8X, and 16X interpolation evaluations on the original `Medium', `Hard', and `Extreme' subsets of SNU-FILM, respectively. 

\noindent\textbf{Evaluation settings.} We calculate PSNR and perceptual indicators, e.g., LPIPS~\cite{zhang2018lpips}, FID~\cite{heusel2017fid}, on both SNU-FILM-arb and X4K-1000FPS (XTest)~\cite{sim2021xvfi}. XTest comprises 15 clips from 15 videos for 8X interpolation. Following the established protocol from prior studies~\cite{hu2022m2m, zhang2023emavfi}, we assess interpolation quality at both 2K and 4K resolutions. For fair comparisons, we disable test-time augmentations during evaluations. We also provide additional evaluations of our method for perception-oriented interpolation in the supplementary materials.

\noindent\textbf{Current advanced VFI methods.} We compare GIMM-VFI with various advanced interpolation methods. For non-INR methods, we include RIFE~\cite{huang2022rife}, IFRNet~\cite{Kong_2022_ifrnet}, M2M~\cite{hu2022m2m}, AMT~\cite{li2023amt}, UPR-Net~\cite{jin2023uprnet}, and EMA-VFI~\cite{zhang2023emavfi}. We also make comparisons with the INR-based method CURE~\cite{shangguan2022cure}.

\noindent\textbf{Results.} 
We present the quantitative results of GIMM-VFI on arbitrary-timestep interpolation benchmarks in Table~\ref{tab:arbres}. Our proposed method achieves high-quality continuous interpolation across various timesteps (i.e., 4X, 8X, 16X). In terms of PSNR, we observe that our RAFT-based~\cite{teed2020raft} method GIMM-VFI-R achieves significant improvements of \textbf{0.18} dB on XTest-2K, \textbf{0.67} dB on XTest-4K, and approximately \textbf{0.30} dB on each subset of the SNU-FILM-arb, in comparison with the previous state-of-the-art method EMA-VFI~\cite{zhang2023emavfi}. 
For perceptual metrics, \textit{i.e.,} LPIPS and FID, our proposed GIMM-VFI still outperforms previous methods on most benchmarks, delivering competitive performance.
These observations demonstrate that our GIMM can offer effective motion modeling for the arbitrary-timestep VFI task. 

\noindent\textbf{Visualizations.} 
Besides the quantitative evaluations, we further qualitatively compare both GIMM-VFI-R and GIMM-VFI-F with existing VFI techniques, as illustrated in Figure~\ref{fig:visnx}. Our methods achieve better interpolation across various timesteps. For example, our methods maintain the integrity of moving object silhouettes ($3{rd}$ case). Moreover, GIMM-VFI-R and GIMM-VFI-F preserve detailed textures within both occluded regions and moving objects from significant deformations, \textit{e.g.,} the rectangle pole tag($1st$ case), truck ($2nd$ case), swing ($3{rd}$ case).

\section{Limitation}
\label{sec:limitations}
There are several known limitations to our method. First of all, the GIMM-VFI is closely related to the pre-trained flow estimator, which estimates bidirectional flows and extracts image features. Therefore, GIMM-VFI inherits the limitations of the chosen pre-trained flow estimators, which may include high computational costs. Additionally, GIMM-VFI only accepts two consecutive frames as inputs, which is not favorable for frame interpolations where larger and nonlinear motion exists.

\section{Conclusion}
We propose a novel motion modeling method, GIMM, which performs continuous motion modeling for video frame interpolation. Our proposed GIMM is the first attempt to learn generalizable implicit neural representations for continuous motion modeling. The method can properly predict arbitrary-timestep optical flows within two adjacent video frames at any resolution, and be easily integrated with existing flow-based VFI approaches (\textit{e.g.,} AMT~\cite{li2023amt}) 
without further modifications by supplying accurately modeled motion.
Extensive experiments on the VFI benchmarks show that our GIMM can effectively perform generalizable motion modeling across videos. 

\noindent\textbf{Acknowledgement.} 
This study is supported under the RIE2020 Industry Alignment Fund – Industry Collaboration Projects (IAF-ICP) Funding Initiative, as well as cash and in-kind contributions from the industry partner(s).


\bibliography{main}

\clearpage
\section{Appendix}
In this supplementary material, we first provide details of our method in Section~\ref{sec:method_details}, including descriptions of flow metrics and reverse flow normalization. Besides, we present the network architectures of both the GIMM module and frame synthesis module in Section~\ref{sec:network_arch}. Moreover, we provide details of our implementation and the corresponding training hyperparameters in Section~\ref{sec:implement} and Section~\ref{sec:hyperparameters} respectively. We show qualitative results of motion modeled by GIMM and its model variants in Section \ref{sec:add_res}. 
Furthermore, we present integrations with other flow-based VFI methods in Section~\ref{sec:integrations}. We also provide an enhanced version of our method for perception-oriented interpolation in Section~\ref{sec:percep_interp}, highlighting the great potential of GIMM. Finally, we discuss the potential broader impacts in Section \ref{sec:impacts}. 

\subsection{Method Details}
\label{sec:method_details}

\noindent\textbf{Flow metrics.} Following Splatting-based Synthesis~\cite{niklaus2023splattingwacv}, the implementation of the forward warping operation within our GIMM module leverages flow metrics to enhance its accuracy. More precisely, we employ flow consistency $U^i_{flow}$, and flow variance $U^i_{var}$. For illustrative purposes, taking $F_{0\rightarrow1}$ as a representative example, we compute its associated metrics as follows:
\begin{align}
    & U^0_{flow} = ||F_{0\rightarrow1} + \overleftarrow{\omega}(F_{1\rightarrow0},F_{0\rightarrow1}) ||_1, \\
    & U^0_{var} = ||\sqrt{G((F_{0\rightarrow1})^2)-G(F_{0\rightarrow1})^2}||,
\end{align}
where $\overleftarrow{\omega}$ is the backward warping operator, and $G(\cdot)$ denotes a $3\times3$ Gaussian filter. 

\noindent\textbf{Reverse flow normalization.}
In GIMM-VFI, we employ reverse flow normalization to convert the predicted normalized flow, $V_t$, into bilateral flows for interpolation purposes. This transformation is expressed through the following equations:
\begin{align}
    & F_{t\rightarrow0} = -t\cdot\phi^{-1}(V_t), \\
    & F_{t\rightarrow1} = (1-t)\cdot\phi^{-1}(V_t).
\end{align}
In this context, $\phi^{-1}$ represents the scaling operation that extends the value range from $[0,1]$ to $\mathbb{R}$ using an instance-specific scaling factor. This reverse flow normalization operation is identical to the one used in IFE~\cite{figueiredo2023frameife}.

\subsection{Network Architecture}
\label{sec:network_arch}
\begin{figure*}[ht]
  \centering
  \includegraphics[width=1.0\linewidth]{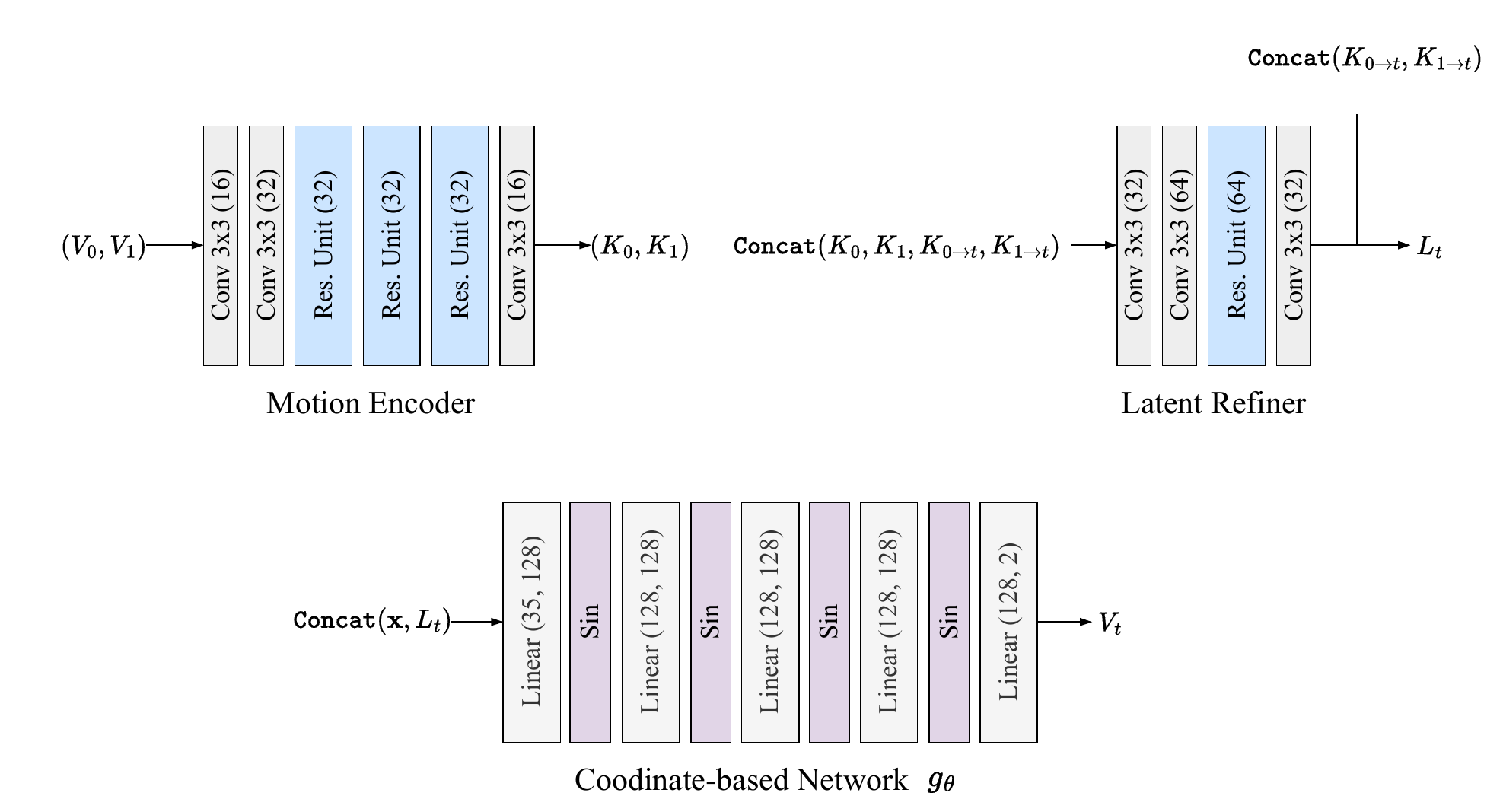}
  \caption{Details of the network architectures within the GIMM Module. The numbers in parentheses denote the output channel count for convolutional layers or the dimensionality shift from input to output for linear layers.}
  \label{fig:gimm_arch}
\end{figure*}

\begin{figure*}[ht]
  \centering
  \includegraphics[width=0.65\linewidth]{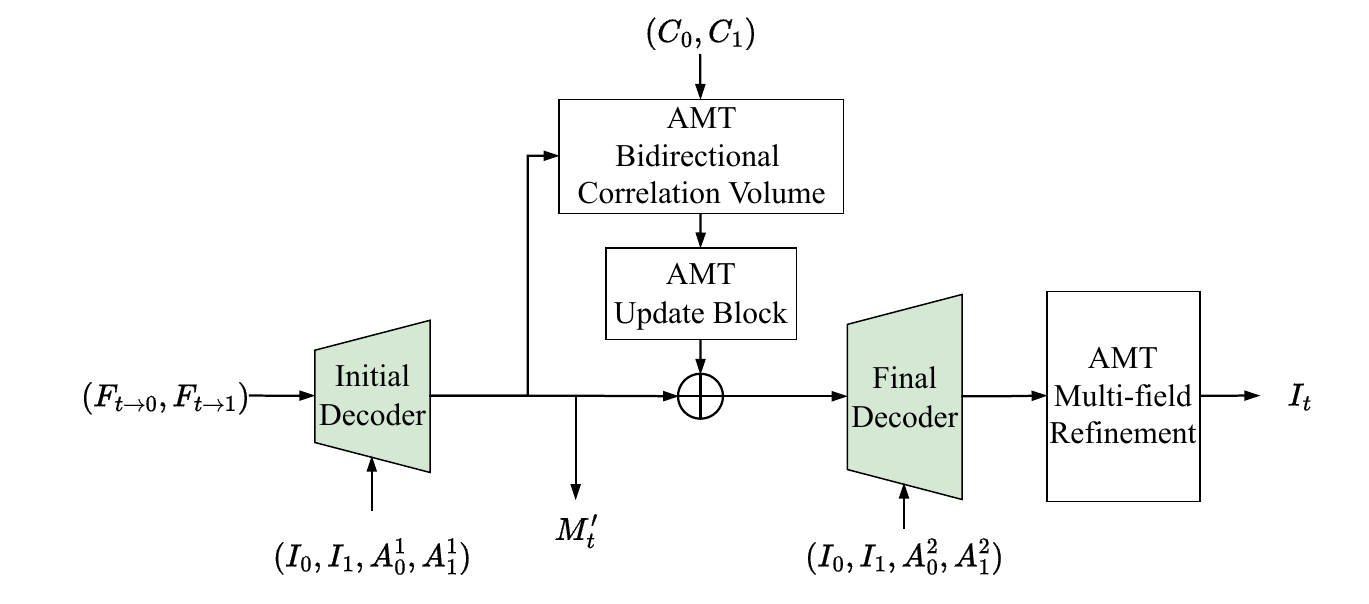}
  \caption{Overview of the frame synthesis module. The frame synthesis module conducts multi-scale flow refinements and predicts warping masks for interpolation, including the intermediate warping mask $M_t'$ predicted by the initial decoder. It integrates the images and their pyramid features ${A_0^l, A_1^l|l\in{1,2}}$ during decoding, updating predictions with correlation information from the Bidirectional Correlation Volume, which is constructed with correlation features $(C_0, C_1)$. Ultimately, we use the outputs from the Final Decoder to perform the final interpolation of $I_t$ through the Multi-field Refinement block. This block is adapted from AMT-G~\cite{li2023amt}. Modified components are highlighted in \textcolor{mgreen}{green} for clarity.}
  \label{fig:fs_arch}
\end{figure*}
\begin{figure*}[ht]
  \centering
  \includegraphics[width=1.0\linewidth]{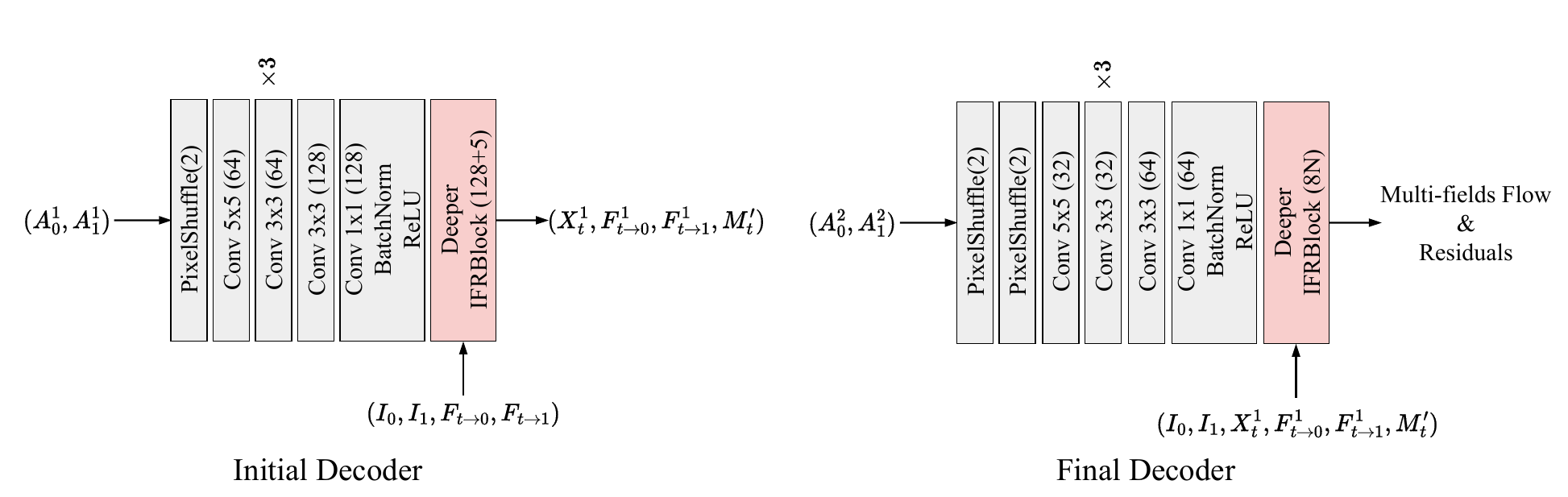}
  \caption{The network architectures of the modified decoders within the frame synthesis module. Unless specifically noted, each convolutional layer is activated by a PReLU function. The Deeper IFRBlock (highlighted in \textcolor{mred}{red}) is based on the decoder architecture (IFRBlock) introduced in IFRNet~\cite{Kong_2022_ifrnet}. In particular, we enhance the original IFRBlock by adding two additional residual blocks to create the Deeper IFRBlock.
}
  \label{fig:dec_arch}
  
\end{figure*}

\noindent\textbf{GIMM module.} 
The architectural details of the networks used in GIMM are shown in Fig. \ref{fig:gimm_arch}. The Motion Encoder and Latent Refiner use a similar convolutional network architecture with residual units. Each Residual Unit consists of two convolutional layers with a skip connection. The major differences between them are the number of channels and residual units. On the other hand, the coordinate-based network $g_\theta$ is designed as a five-layer SIREN~\cite{sitzmann2020siren}, incorporating the motion latent $L_t$ as an auxiliary input. 

\noindent\textbf{Frame synthesis module.} 
Adapted from AMT-G~\cite{li2023amt}, the framework of the Frame synthesis module is illustrated in Fig. \ref{fig:fs_arch}. This module integrates two unique decoding stages: an adapted initial decoder and a revised final decoder, while the rest of the module's architecture remains unchanged from AMT-G. The architectural specifics of these modified decoders are shown in Fig. \ref{fig:dec_arch}. 

\subsection{Implementation Details}
\label{sec:implement}
In addition to experiments in the main text, we present more detailed implementations of GIMM and GIMM-VFI in this section.

\noindent\textbf{GIMM training.} We begin by extracting optical flows from the training split of the Vimeo90K triplets dataset~\cite{xue2019vimeo90k} using Flowformer~\cite{huang2022flowformer}. With these extracted flows, we train GIMM, randomly cropping the flows to a resolution of $256 \times 256$. For each batch during training, we randomly select a timestep $t$ from the set $\{0, 0.5, 1\}$ to supervise. We set the batch size to 64, and train the model for 240 epochs with a learning rate of $1 \times 10^{-4}$, using 2 NVIDIA V100 GPUs. 

\noindent\textbf{GIMM-VFI training.} We integrate the pre-trained GIMM module and flow estimator into the GIMM-VFI model for end-to-end training on the arbitrary-timestep interpolation task. We implement two variants of GIMM-VFI, using two different flow estimators: the RAFT~\cite{teed2020raft} and FlowFormer~\cite{huang2022flowformer}, designated as GIMM-VFI-R and GIMM-VFI-F, respectively. However, both versions of GIMM-VFI share the same training process. Similar to previous works~\cite{zhang2023emavfi,huang2022rife}, we train our model on the complete Vimeo90K septuplet split~\cite{xue2019vimeo90k} for 60 epochs with a batch size of 32 and a learning rate of $8 \times 10^{-5}$. We randomly select triple subsets for training from each septuplet, following the same sampling strategy as previous research~\cite{zhang2023emavfi, huang2022rife}. We resize and randomly crop each frame into a resolution of $224\times224$ and perform a series of augmentations including rotation, flipping, temporal order reversing and channel order reversing. We train our model on 8 NVIDIA V100 GPUs.

\subsection{Training hyperparameters}
\label{sec:hyperparameters}

In addition to the implementation details in Sec.~\ref{sec:implement}, we summarize the training settings for both GIMM and GIMM-VFI in Tab. \ref{tab:gimmvfi}.

\begin{table}[ht]
\centering
\caption{\textbf{Training settings} for GIMM and GIMM-VFI.}
\resizebox{0.55\textwidth}{!}{
\begin{tabular}{lcc}
\toprule
Configuration & GIMM   & GIMM-VFI \\
\midrule
Optimizer & \multicolumn{2}{c}{AdamW} \\
Peak learning rate & 1e-4 & 8e-5 \\
Minimum learning rate & 1e-4 & 8e-6 \\
Epochs & 240 & 60 \\
Batch size per GPU & 16 & 4 \\
Weight decay & 0 & 4e-5 \\
Optimizer momentum & \multicolumn{2}{c}{$\beta_1, \beta_2 =$ 0.9, 0.999} \\
Learning rate schedule & \xmark & Cosine annealing \\
Warmip epochs & 0 & 1 \\
Training Resolution & $256\times256$ & $224\times224$ \\
Num. V100 GPUs & 2& 8 \\
\bottomrule
\end{tabular}
}
\label{tab:gimmvfi}
\end{table}

\subsection{Qualitative Results of GIMM Motion Modeling}
\label{sec:add_res}
\begin{figure*}[ht]
  \centering
  \includegraphics[width=1.0\linewidth]{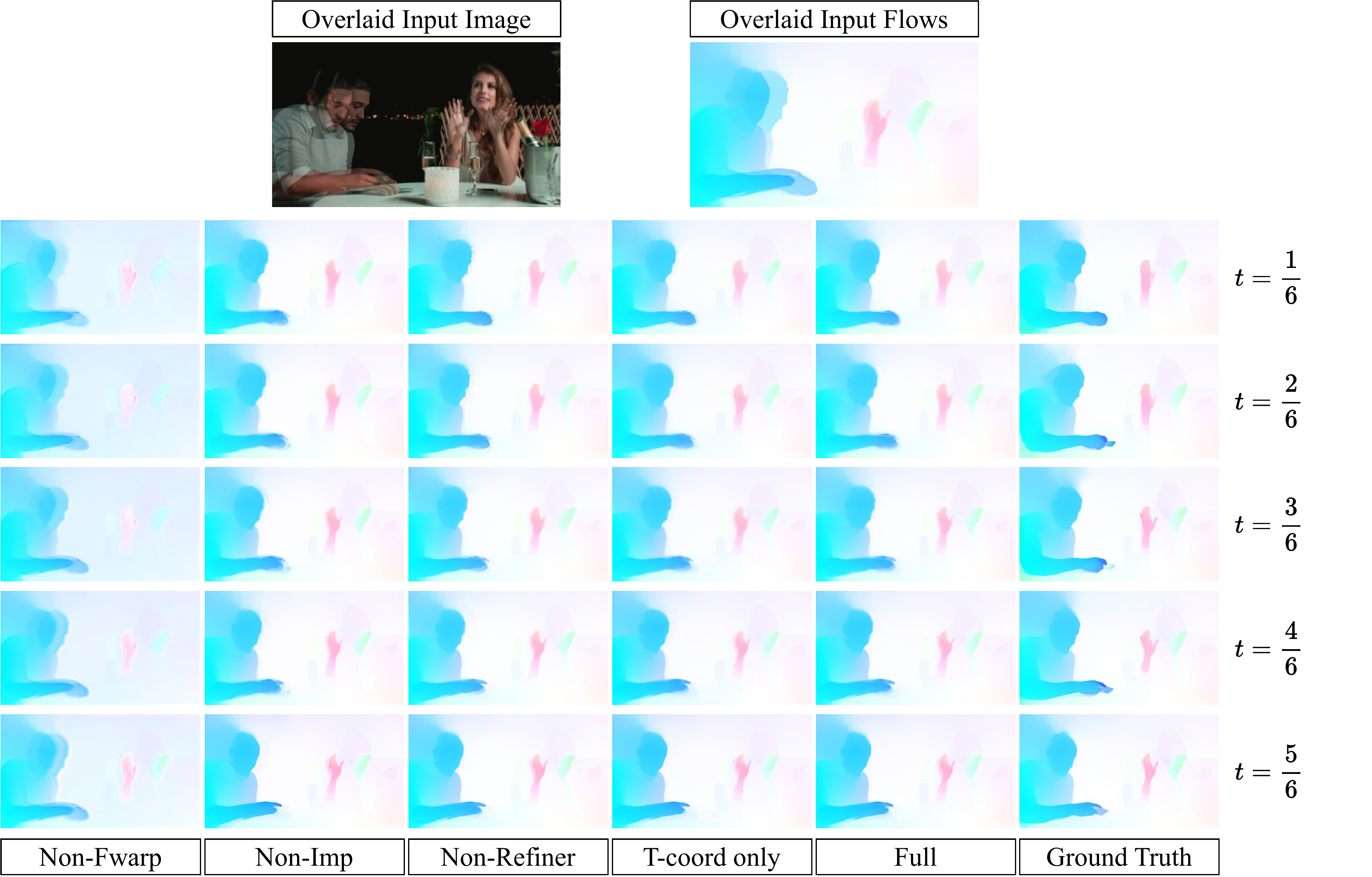}
  \caption{Visual comparisons for 6X motion modeling on the Vimeo-septuplet-flow. We compare GIMM with its model variants mentioned in the ablation study of the main text.}
  \label{fig:supp_flow}
\end{figure*}

We present additional qualitative results of GIMM motion modeling in Figure~\ref{fig:supp_flow}. The visualization showcases the modeled motion of GIMM and variants in the ablation study. Our observation confirms the effectiveness of GIMM's design and is consistent with the results from the ablation study in the main text. For example, the head motion in Figure~\ref{fig:supp_flow} significantly deteriorates with noises across all timesteps when forward warping is replaced or latent refinement is skipped.

\subsection{Integration with other flow-based VFI methods}
\label{sec:integrations}
Our method GIMM focuses on continuous motion modeling, which further enables frame interpolation at arbitrary timesteps. We additionally integrate GIMM with other existing flow-based VFI approaches (\textit{e.g.}, TTVFI~\cite{liu2023ttvfi}, IFRNet~\cite{Kong_2022_ifrnet}), by supplying accurately modeled motion. For evaluations, we calculate PSNRs on arbitrary-timestep interpolation benchmark SNU-FILM-arb in Table~\ref{tab:integration_GIMM}. We observed that integrating the GIMM module resulted in significant improvements. This further demonstrates the effectiveness of GIMM for continuous motion modeling when integrated with existing flow-based VFI works.

\begin{table}[!t]
    \caption{\textbf{Integrating GIMM with TTVFI and IFRNet.} 
    }
    \centering
    \begin{tabular}{lccc}
    \toprule
    \multirow{2}{*}{Motion method} & SNU-FILM-arb-Medium & SNU-FILM-arb-Hard & SNU-FILM-arb-Extreme\\
    \cmidrule(lr){2-4}
     & PSNR$\uparrow$ & PSNR$\uparrow$ & PSNR$\uparrow$ \\
    \midrule
    TTVFI & 34.48 & 30.39 & 26.24 \\
    TTVFI+GIMM & \textbf{35.55 (+1.07dB)} & \textbf{31.60 (+1.21dB)} & \textbf{27.40 (+1.16dB)} \\
    IFRNet & 34.88 & 31.15 & 26.32 \\
    IFRNet+GIMM & \textbf{36.46 (+1.58dB)} & \textbf{32.20 (+1.05dB)} & \textbf{27.73 (+1.41dB)} \\
    \bottomrule
    \end{tabular}
    \label{tab:integration_GIMM}
    
\end{table}

\subsection{Perception-oriented interpolation}
\label{sec:percep_interp}
Our proposed method can be further enhanced for perception-oriented interpolation by incorporating an additional learning objective, specifically the LPIPS~\cite{zhang2018lpips} loss. The perceptually enhanced GIMM-VFI variants are denoted as GIMM-VFI-R-P and GIMM-VFI-F-P respectively.
For evaluation, we compute the LPIPS and FID indicators on the same benchmarks as those used in Section~\ref{sec:interpo} of the main paper.

\noindent\textbf{Results.} We present enhanced performance as measured by perceptual metrics in Table~\ref{tab:arbres_percep}. The enhanced variants, GIMM-VFI-R-P and GIMM-VFI-F-P, demonstrate significant performance improvements, particularly on the Extreme (16X) subset of the SNU-FILM-arb dataset, where LPIPS and FID scores decrease by approximately 0.05 and 9, respectively. These results underscore the effectiveness of our enhancement strategy and highlight the substantial potential of the GIMM module in delivering accurate motion modeling for arbitrary-timestep interpolation.

\begin{table*}[!t]
    \caption{\textbf{Quantitative improvements for perception-oriented arbitrary-timestep interpolation.}
 We report the perceptual indicators as LPIPS↓/ FID↓, with the best results highlighted in \sota{boldface} and the second best results in \subsota{underline}. 
    }
    \vspace{5pt}
    \centering
    \resizebox{\textwidth}{!}{
    \begin{tabular}{lccccc}
\toprule
         \multirow{2}{*}{Method} & \multicolumn{2}{c}{XTest}& \multicolumn{3}{c}{SNU-FILM-arb} \\
         \cmidrule(lr){2-3} \cmidrule(lr){4-6} 
         &2K & 4K & Medium (4X) & Hard (8X) & Extreme (16X) \\
\midrule
        GIMM-VFI-R &0.113~/~6.52 & 
        0.149~/~6.49 & 
        0.033~/~5.89 & 
        0.060~/~9.59 & 
        0.110~/~16.45\\
        GIMM-VFI-F &0.103~/~6.74 & 
        0.142~/~6.58 & 
        0.031~/~5.86 & 
        0.059~/~9.95 & 
        0.109~/~16.79\\
\midrule
        GIMM-VFI-R-P &\sota{0.047}~/~4.62 & 
        0.099~/~4.74 & 
        \sota{0.016}~/~3.81 & 
        \sota{0.030}~/~\sota{4.93}	& 
        0.059~/~\sota{7.30}\\
        GIMM-VFI-F-P &
        \sota{0.047}~/~\sota{4.42} & 
        \sota{0.098}~/~\sota{4.57} &
        0.017~/~\sota{3.76} & 
        \sota{0.030}~/~\sota{4.93}	 & 
        \sota{0.058}~/~7.40\\
\bottomrule
    \end{tabular}
    }
    \label{tab:arbres_percep}
    
\end{table*}

\subsection{Broader Impacts}
\label{sec:impacts}

Our proposed method, GIMM-VFI, demonstrates strong performance in continuous frame interpolation, with applications spanning real-world scenarios such as creating cinematic special effects and producing slow-motion videos. This technique not only enhances these use cases but also opens doors for further research in video technology. At its core, GIMM-VFI employs the GIMM module for continuous motion modeling in video frame interpolation, a feature that could also be adapted for related tasks like video compression and predictive video analysis. Despite its strengths, our approach shares limitations with other frame interpolation methods and, in some cases, may introduce artifacts that could pollute visual data on the Internet.

\clearpage

\end{document}